\documentclass{article}

\usepackage{iclr2027_conference,times}

\usepackage{amsmath,amsfonts,bm}

\def\eqref#1{equation~\ref{#1}}

\def\1{\bm{1}}

\DeclareMathAlphabet{\mathsfit}{\encodingdefault}{\sfdefault}{m}{sl}
\SetMathAlphabet{\mathsfit}{bold}{\encodingdefault}{\sfdefault}{bx}{n}

\usepackage[utf8]{inputenc}
\usepackage[T1]{fontenc}

\usepackage{hyperref}
\usepackage{url}
\usepackage{booktabs}
\usepackage{amsfonts}
\usepackage{amsmath}
\usepackage{amssymb}
\usepackage{nicefrac}
\usepackage{microtype}
\usepackage{xcolor}
\usepackage{graphicx}
\usepackage{placeins}

\usepackage{xcolor}

\usepackage{enumitem}

\newcommand{\Eq}[1]{Eq.~(\ref{eq:#1})}

\newcommand{\fig}[1]{Fig.~\ref{fig:#1}}

\usepackage{amsmath}

\usepackage{xspace}

\title{The RNN Homunculus: \\ Learning to Gate Long-Term Memory}

\author{Zeyu Fan, Alexandru-Dragos Manolache \& Jan van Gemert \\
Delft University of Technology\\
Mekelweg 5, 2628 CD Delft, The Netherlands\\
}

\iclrfinalcopy

\begin{document}

\maketitle
\fancyhead{}
\renewcommand{\headrulewidth}{0pt}

\begin{abstract}
Learning long-range dependencies with recurrent neural networks (RNNs) is difficult because of the vanishing gradient problem. Gated recurrent neural networks like GRUs and LSTMs address this problem and are commonly explained by their ability to create additive copy paths through time. This explanation is correct, but incomplete: useful RNN gate values must themselves be learned, which is typically done with another RNN. We study this gate-learning problem through carefully controlled sequence classification tasks. Our analysis shows that gated architectures struggle when all training samples require long-range memory and thus equally suffer from the vanishing gradient problem. We intervene by adding short-dependency samples which let gated models learn solutions that capture long-dependency samples, but also generalize across different class labels.
When properly learned, gated RNNs match or even exceed the test accuracy of modern architectures like Mamba and xLSTM. Overall, our results highlight gate learning as a central part of learning long-range memory and show how training data can support this process by exposing the same task over shorter dependencies. 
\end{abstract}

\section{Introduction}
Long-range sequence modeling is important for language models, chatbots, and agentic AI~\citep{brown2020language,wang2024survey}. However, modern sequence models face a tension between long-range reasoning and inference cost.
Self-Attention has become the dominant mechanism for modeling long sequences, but its inference cost grows quadratically with sequence length~\citep{vaswani2017attention}.
Recurrent Neural Networks (RNNs) may offer a complementary trade-off: they update fixed-size states recurrently, leading to linear-time sequence processing during inference~\citep{elman1990finding,williams1989learning}.
The obstacle is that vanilla RNNs struggle to learn long-range dependencies. Backpropagation through time
multiplies many Jacobians, which can cause the gradients to vanish as the sequence length grows~\citep{bengio1994learning,hochreiter1998vanishing,pascanu2013difficulty}.
Understanding how recurrent models can overcome the vanishing gradient problem and long-range credit assignment
remains relevant because state-space architectures
continue to be attractive for efficient long-range modeling~\citep{gu2022efficiently}.

\begin{figure}[t]
  \centering
  \includegraphics[width=0.95\linewidth]{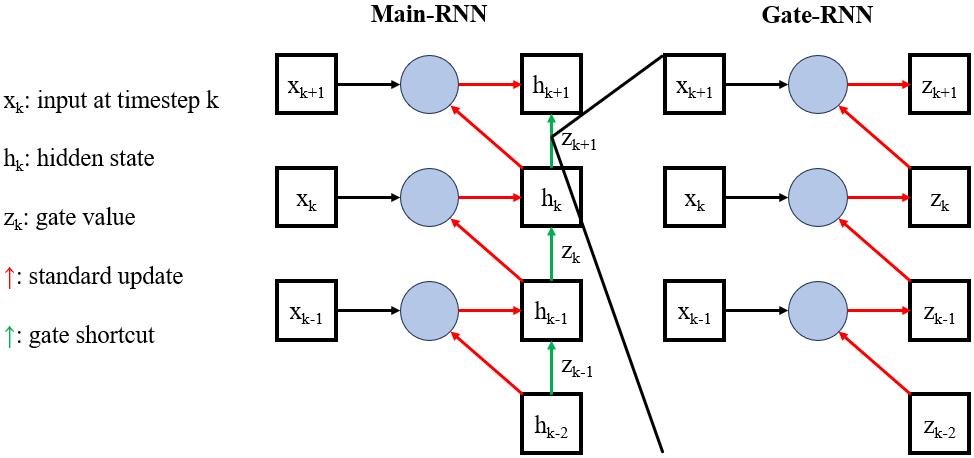}
  \caption{\textbf{The homunculus problem in gated RNNs.} 
  The left side shows a gated RNN: the standard recurrent computation, shown in red, updates the hidden state \(h_k\) with the previous hidden state and the current input \(x_k\). This path forms the long chain through which gradients can vanish. Gate values \(z_k\) create direct paths, shown in green, directly from \(h_{k-1}\) to \(h_k\).
  The zoom-in on the right shows the missing step in this explanation: the gate values \(z_k\) are themselves computed from inputs \(x_k\) and previous states by a recurrent process, a "gate-RNN" with identical structure to the main gated RNN.
  }
  \label{fig:rnn-homunculus}
\end{figure}

Gated recurrent networks, such as Long Short-Term Memory (LSTM)~\citep{hochreiter1997long} and Gated Recurrent Units (GRUs)~\citep{cho2014learning} were developed to address the vanishing gradient problem in RNNs~\citep{chung2014empirical}.
Their hidden state updates contain additive paths controlled by gates, which can create near-identity routes for information and gradients.
For example, an LSTM cell can preserve information when the forget gate remains close to 1, and a GRU can preserve information when its update gate keeps the previous hidden state~\citep{noh2021analysis}.
This mathematical explanation is correct, but incomplete. There is a \emph{homunculus} in it.
A homunculus explanation assigns a difficult behavior to an internal mechanism that requires the same explanation, producing a regress rather than resolving the original difficulty~\citep{kenny1971homunculus,ryle2000concept,dennett1991consciousness}.
The standard explanation of gated RNNs assumes useful gate values, but the gates are themselves learned functions of the current input and previous hidden state.
Therefore, the gate values that protect long-range gradients must also be learned through the same recurrent computation that suffers from long-range gradient decay.
~\fig{rnn-homunculus} illustrates this regress of recurrent structures from a ``Main-RNN'' to a ``Gate-RNN''.

This paper studies the missing learning problem of gates. 
The central question is how gated RNNs learn the gate configurations that support long-range memory.
Our analysis challenges the current popular belief, and shows that gates do not inherently solve the gradient problem and long-range dependencies.
We hypothesize that gated RNNs succeed when the data contains a short-range sub-task that can be learned before the full long-range task is solved.
Once the Gate-RNN has learned the sub-task, the Main-RNN can use the induced memory path to solve long-range tasks.

We test the hypotheses with controlled sequence classification tasks, designed as a temporal XOR variant in which two informative tokens determine the label amid non-informative tokens.
When all training samples require long-range memory, vanilla RNNs, GRUs, and LSTMs all fail as the sequence length increases.
We intervene by adding short-dependency samples in which the same label rule can be learned from information separated by shorter temporal gaps and thus
which allow the Gate-RNN to learn and enable gated RNNs to solve the original long-range task. Because gradient flow through the Gate-RNNs benefits all classes in the Main-RNN we evaluate how gate learning is shared over classes. Specifically, we evaluate if gate learning in one set of classes generalizes to  different classes. We explore modern gated RNNs and show that short-dependency training support also benefits gates that depend only on the current input.
With this support, gated RNNs match or exceed the test accuracy of Mamba~\citep{gu2023mamba} and xLSTM~\citep{beck2024xlstm}.

Our contributions can be summarized as follows:
\begin{itemize}
    \item We identify a missing step in the standard explanation of gated RNNs: gates can create good gradient paths after useful gate values have been learned, but gate learning itself is also subject to recurrent gradient decay.
    \item We introduce controlled experiments that demonstrate when gated RNNs fail and succeed. We show that short-dependency samples enable learning that generalizes to long-range.
    \item We show that gates generalize over classes, where gate learning in one class benefits long-range dependency learning in different classes. 
    \item We show that short-dependency training support enables gated RNNs to close the accuracy gap with Mamba and xLSTM.
\end{itemize}

\section{Related Work}
\paragraph{The Vanishing Gradient Problem} 
Learning long-range dependencies with gradient descent requires signals from a later loss to pass through many recurrent transitions~\citep{bengio1994learning,hochreiter1998vanishing,pascanu2013difficulty}.
Classical analyses express this propagation as a product of recurrent Jacobians, which can decay or grow exponentially with temporal distance.
When these products decay, long-range contributions to learning can become negligible relative to short-range ones.
This analysis motivates architectures with additive state paths that can preserve information and gradients across longer intervals.

\paragraph{Gated Recurrent Networks}
LSTMs introduce memory cells with multiplicative gates to control the storage and retrieval of information while supporting constant gradient flow~\citep{hochreiter1997long}.
GRUs simplify gated recurrence through reset and update gates~\citep{cho2014learning}.
Empirical comparisons show advantages over vanilla RNNs, with the relative accuracy of GRUs and LSTMs varying across tasks~\citep{chung2014empirical,jozefowicz2015empirical}.
Both architectures provide additive copy paths controlled by learned gates.
The LSTM forget gate and GRU update gate control how much of the previous state is retained along these paths.
Their ability to preserve information and gradients therefore depends on gate values that need to be learned as well.

\paragraph{Modern Recurrent Architectures}
Recent methods revisit recurrent sequence modeling through simpler gates or richer memory updates.
minGRU and minLSTM~\citep{feng2024were} simplify gated recurrence by computing both their gates and proposed state updates from the current input alone.
Mamba~\citep{gu2023mamba} makes selected state-space parameters input-dependent, allowing the model to retain or forget information while scaling linearly with sequence length.
xLSTM~\citep{beck2024xlstm} extends LSTM with exponential gating, normalization, and modified memory structures.
Its sLSTM variant uses scalar memory with memory mixing, while mLSTM uses matrix memory and supports parallel computation.
Our experiments compare Mamba and xLSTM with gated RNNs to examine how short-dependency samples affect their accuracy on long-dependency tasks.

\paragraph{What Gated Recurrent Networks Learn}
Studies of trained LSTMs reveal learned long-range structure and its limitations.
In subject--verb agreement, targeted supervision produces strong results, but conflicting linear cues can reduce reliability~\citep{linzen2016assessing}.
LSTM language models also capture long-distance agreement across languages~\citep{gulordava2018colorless}, and track filler--gap dependencies across positions and constructions~\citep{kobzeva2022lstms}.
Mechanistic analyses identify cells that track quotes, brackets, and line length~\citep{karpathy2015visualizing}, as well as units that maintain grammatical number and others that regulate its storage and release~\citep{lakretz2019emergence}. Our work complements these studies by examining how training data helps gated RNNs learn to retain task-relevant information across time.
We analyze learning signals reaching the gates and test how short-dependency samples support long-range learning.

\section{The Gate Learning Problem}
\label{sec:gate-learning}
\label{sec:gate-learning-problem}


Consider a GRU-style gated update~\citep{cho2014learning},
\begin{equation}
h_t = z_t \odot h_{t-1} + (1-z_t)\odot \tilde{h}_t ,
\label{eq:gru-style-update}
\end{equation}
where $h_t \in \mathbb{R}^{d_h}$ is the hidden state at sequence position $t$, $\tilde{h}_t \in \mathbb{R}^{d_h}$ is a candidate state, $z_t \in [0,1]^{d_h}$ is the update gate, and $\odot$ denotes element-wise multiplication. 
With this convention, a gate value close to $1$ copies the previous state, while a value close to 0 writes the candidate state.

Let $L$ be a loss that depends on the final hidden state $h_T$, and let $\delta_t = \partial L / \partial h_t$ denote the gradient arriving at hidden state $h_t$. 
Differentiating \Eq{gru-style-update}
with respect to the Gate-RNN's gate activation, holding its other inputs fixed, gives
\begin{equation}
\frac{\partial L}{\partial z_t}
=
\delta_t \odot (h_{t-1}-\tilde{h}_t).
\label{eq:gate-activation-gradient}
\end{equation}
This gradient tells the model whether increasing the copy path or the write path would reduce the loss at sequence position $t$. 
However, the strength of this signal depends directly on $\delta_t$.

For a final-step loss, $\delta_t$ is obtained by backpropagating through all future recurrent transitions:
\begin{equation}
\delta_t
=
J_{t+1}^{\top}J_{t+2}^{\top}\cdots J_T^{\top}\delta_T,
\qquad
J_i = \frac{\partial h_i}{\partial h_{i-1}} .
\label{eq:hidden-gradient-chain}
\end{equation}
Here $J_i$ is the full recurrent Jacobian, including the dependence of the gate and candidate on the previous state when present.
Thus, the learning signal for an early gate also passes through a long product of recurrent Jacobians. In particular,
\begin{equation}
\left\|\frac{\partial L}{\partial z_t}\right\|_2
\leq
\|h_{t-1}-\tilde h_t\|_\infty
\left\|J_{t+1}^{\top}\cdots J_T^{\top}\right\|_2
\|\delta_T\|_2.
\label{eq:gate-gradient-bound}
\end{equation}
If the Jacobian product contracts while the other factors remain bounded, the gradient reaching the early gate is attenuated.
Gates may therefore receive small learning signals.
We next examine how this signal reaches the parameters that compute the gate values.

For a conventional state-dependent sigmoid gate,
\begin{equation}
z_t = \sigma(a_t^z),
\qquad
a_t^z = W_z x_t + U_z h_{t-1} + b_z ,
\label{eq:gate-param}
\end{equation}
where $x_t$ is the input, $W_z$, $U_z$, and $b_z$ are gate parameters, and $\sigma$ is the logistic sigmoid.
The gradient with respect to its preactivation is
\begin{equation}
\frac{\partial L}{\partial a_t^z}
=
\frac{\partial L}{\partial z_t}
\odot z_t(1-z_t).
\label{eq:gate-preactivation-gradient}
\end{equation}
A small incoming gradient therefore limits the local gate-learning signal, and sigmoid saturation can attenuate it further.

The same issue appears in LSTMs.
For the cell update~\citep{hochreiter1997long}
\begin{equation}
c_t = f_t \odot c_{t-1} + i_t \odot \tilde{c}_t,
\label{eq:lstm-cell-update}
\end{equation}
the forget gate receives
\begin{equation}
\frac{\partial L}{\partial f_t}
=
\frac{\partial L}{\partial c_t}
\odot c_{t-1}.
\label{eq:lstm-forget-gradient}
\end{equation}
A forget gate close to $1$ provides a direct path for preserving cell-state gradients, but its learning signal still depends on the gradient arriving at $c_t$.
For both GRUs and LSTMs, learning useful gate values therefore depends on the propagation of task-loss gradients through the recurrent computation. 


\subsection{Extension to More General Gating}
\label{sec:general-gates}

The gate-gradient analysis also applies to more general gated recurrent updates:
minGRU and minLSTM~\citep{feng2024were} compute gates and candidate states from the current input alone.
We examine whether removing hidden states dependency also removes the gate-learning difficulty.

The minGRU update and the normalized minLSTM update can both be written as \Eq{gru-style-update}.
For minGRU, $z_t$ is the complement of its write gate.
For normalized minLSTM, $z_t=f_t/(f_t+i_t)$, where $f_t$ and $i_t$ are its forget and input gates.
The computations are $z_t=q_\theta(x_t)$ and $\tilde h_t=v_\phi(x_t)$, where $q_\theta$ computes the gate values and $v_\phi$ computes the candidate state, with parameters $\theta$ and $\phi$.
Neither computation depends on $h_{t-1}$ now.

For a fixed input, the previous state enters the update only through $z_t\odot h_{t-1}$. The recurrent Jacobian then simplifies to
\begin{equation}
J_t=\operatorname{diag}(z_t),
\qquad
\delta_t=
\delta_T\odot\prod_{k=t+1}^{T}z_k,
\label{eq:input-only-gradient-chain}
\end{equation}
where the product is element-wise. Substituting into ~\Eq{gate-activation-gradient} gives
\begin{equation}
\frac{\partial L}{\partial z_t}
=
(h_{t-1}-\tilde h_t)\odot\delta_T
\odot\prod_{k=t+1}^{T}z_k.
\label{eq:input-only-gate-gradient}
\end{equation}

For any coordinate $j$, if $0\leq z_{k,j}\leq\rho<1$ for all $k>t$, then
$\prod_{k=t+1}^{T}z_{k,j}\leq\rho^{T-t}$.
Thus, the contribution of the final loss to the gate-learning signal can decrease exponentially with temporal distance.
Computing gates from the current input alone therefore does not guarantee strong learning signals for gates at early positions. Removing previous-state dependence simplifies the recurrence, but does not solve the gate-learning problem.

\subsection{Learning from Short Dependencies}
\label{sec:gate-learning-implications}

Gate parameters are shared across sequence positions. Learning signals from different positions therefore contribute to the same parameter update.
For the gates' weight matrix $W_z$ in ~\Eq{gate-param}, the gradient is
\begin{equation}
\frac{\partial L}{\partial W_z}
=
\sum_{t=1}^{T}
\frac{\partial L}{\partial a_t^z}x_t^\top.
\label{eq:shared-gate-gradient}
\end{equation}
An update learned at one position changes the gates used throughout the sequence.
In short-dependency samples whose task-relevant inputs appear near the final loss, gate-learning signals pass through fewer transformations, reducing the number of factors that can attenuate them.
We hypothesize that these samples help the gates learn which inputs to write into memory and which to ignore. The same gates are then applied to long-dependency samples, where selective updates must preserve information across longer intervals.

\section{Experiments}
\label{sec:experiments}
We carefully design controlled tasks, inspired by a temporal XOR, to examine how short-dependency training samples affect learning over long dependencies.
Each sequence contains two informative tokens that determine its label, with all remaining positions filled by non-informative tokens.
This setup lets us vary the positions of task-relevant information while keeping the label rule and sequence length unchanged.

\begin{figure}[t]
    \centering
    \includegraphics[width=\linewidth]{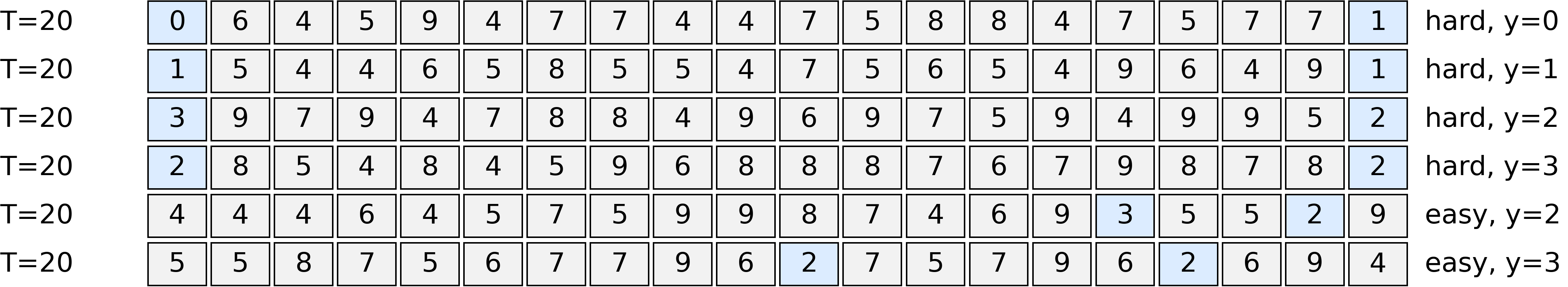}
    \caption{
    \textbf{Multi-class temporal XOR with long and short dependencies.}
    Each row is a length-$20$ sequence. Blue cells mark the two informative tokens; grey cells contain non-informative tokens.
    Labels depend on the informative-token group and whether the two tokens are equal: group $\{0,1\}$ uses labels $0$ (unequal) and $1$ (equal), while group $\{2,3\}$ uses labels $2$ (unequal) and $3$ (equal).
    The first four rows show long-dependency (hard) samples with informative tokens at the sequence endpoints. The last two rows show short-dependency (easy)
    samples from group $\{2,3\}$, with both informative tokens inside the final ten positions.
    Moving these tokens closer to the final prediction provides shorter dependencies for learning the same classification rule.
    }
    \label{fig:mxor-examples}
\end{figure}

\paragraph{Tasks.}
~\fig{mxor-examples} illustrates multi-class temporal XOR: the label identifies which informative-token group is present and whether its two tokens are equal.
For example, informative tokens $0$ and $1$ receive label $0$, while two $0$s receive label $1$.
Consider two informative-token groups, $\{0,1\}$ and $\{2,3\}$.
Within the first group, unequal tokens receive label $0$ and equal tokens label $1$; within the second group, unequal tokens receive label $2$ and equal tokens label $3$.
The eight possible informative-token pairs are balanced, yielding four balanced classes.
Non-informative tokens are sampled uniformly from $\{4,\ldots,11\}$.
The two groups allow us to test whether training support provided to one group also benefits the other.
Four-class chance accuracy is $25\%$; identifying the group from one given informative token and guessing between its two labels yields $50\%$.

The long-dependency samples place the informative tokens between a length-$T$ sequence. We also refer to these as \emph{hard samples}, because solving them requires carrying information across the full recurrent chain.
Short-dependency \emph{easy samples}, place informative tokens at two distinct, randomly chosen positions within the final $\min(10,T)$ positions, see~\fig{mxor-examples}.
Both sample types have the same total length and follow the same label rule.

\paragraph{Models.}
We evaluate recurrent models where each token is mapped to a learnable embedding and classified using its final hidden state.
GRUs and LSTMs use a hidden size of $10$.
The vanilla RNN uses hidden size $10$ when trained only on hard samples, and $24$ when trained also on short-dependency samples. This gives the vanilla RNN a larger recurrent state and a comparable number of parameters
to the gated RNNs. Model configurations and optimization details are provided in Appendix~\ref{app:ExpA}.

\paragraph{Evaluation.}
For each sequence length, we generate separate training, validation, and test sets and train each model from scratch.
Sequences are unique within and across these splits.
Each class has $1{,}000$ training, $200$ validation, and $200$ test samples.
All validation and test samples use hard-only placement.
We select the checkpoint with the highest validation accuracy, using validation loss to break ties, and report its test accuracy.
The experiments use all combinations of five model-initialization seeds and five data-generation seeds ($25$ runs per setting).
We report means and standard deviations across runs, together with group-wise test accuracies for multi-class temporal XOR.

\subsection{How do Gates Learn Long Range Dependencies?}
\label{sec:learning-long-dependencies}

We first test whether gated RNNs also fail at long-range and whether short-dependency samples help gated RNNs learn the hard multi-class XOR task.
We compare hard-only training with mixed training, where $20\%$ of samples in each informative-token group use short-dependency placement.

\begin{figure}[t]
    \centering
    \includegraphics[width=\linewidth]{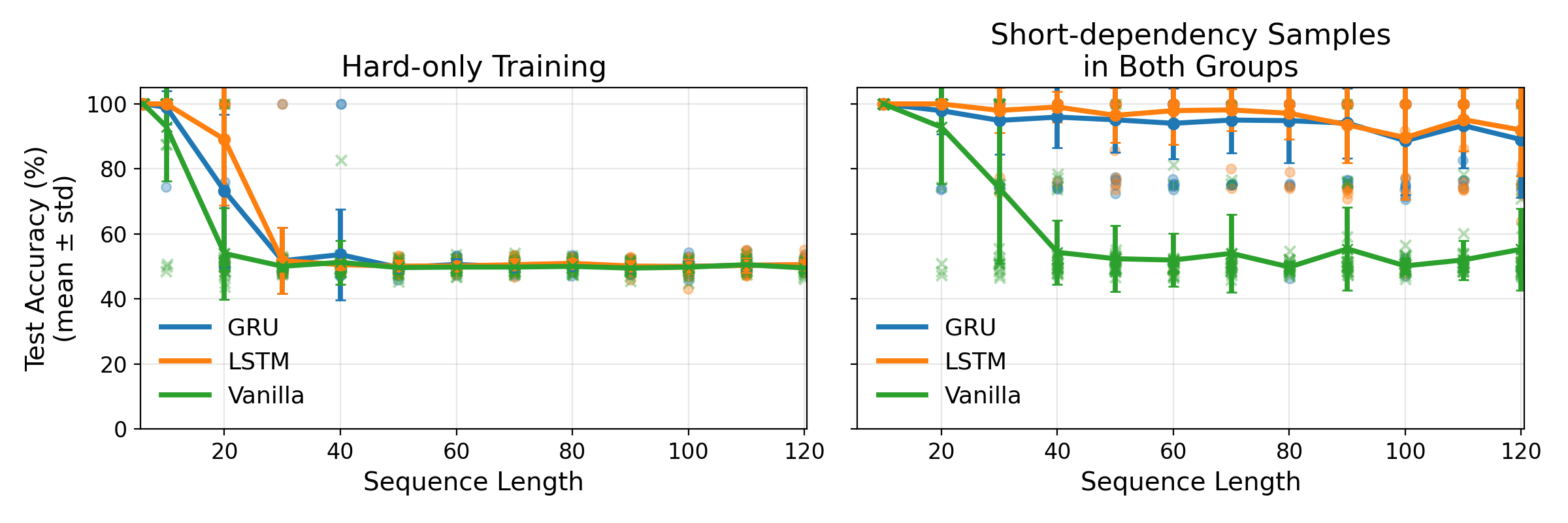}
    \caption{
    \textbf{Short-dependency training samples improve long-range
    generalization in gated RNNs.}
    Test accuracy on multi-class temporal XOR is plotted against total sequence length.
    The task has four labels, determined by the informative-token group and whether its two tokens are equal.
    Left: all training samples are hard samples. 
    Right: for all classes, $20\%$ of training samples have short dependencies, with both informative tokens placed within the final ten positions.
    Both conditions are evaluated exclusively on hard samples.
    Lines show means over $25$ runs, error bars indicate standard deviations, and faint markers show individual runs.
    Four-class chance accuracy is $25\%$; identifying the token group and guessing between its two labels yields $50\%$.
    GRUs and LSTMs fall to approximately $50\%$ on long sequences under hard-only training, but maintain high accuracy with short-dependency samples, while vanilla RNNs remain near $50\%$.
    Thus, short-dependency training samples  enable gated models to generalize to long dependencies that they struggle to learn directly.
    }
    \label{fig:mxor-training-comparison}
\end{figure}

~\fig{mxor-training-comparison} shows that all three architectures learn the task at short sequence lengths.
Under hard-only training, as the sequences grow longer, the vanilla RNN deteriorates first.
GRUs and LSTMs also drop to $50\%$ on long sequences. 
The gated architectures therefore struggle to learn the long-range computation required by the task, showing the gate-learning problem.

With short-dependency samples included in training, GRUs and LSTMs maintain mean test accuracies of approximately $90\%$ or higher across the tested lengths. 
The vanilla RNN stays close to its hard-only accuracy on long sequences, despite using a larger hidden state in mixed training ($24$ units rather than $10$).
The improvement in gated models is measured entirely on hard test samples, showing that the benefit extends beyond the short-dependency examples encountered during training.

These results support the gate-learning hypothesis developed in Section~\ref{sec:gate-learning-implications}.
Short-dependency samples place informative inputs closer to the final loss, providing shorter paths for learning when to write information and when to preserve memory.
The shared gate functions can then apply this behavior across longer intervals.

\subsection{Does Gate learning Transfer Across Classes? 
}
\label{sec:transfer-across-groups}

Does the benefit of short-dependency training extend beyond the tokens and labels receiving this support?
We now provide short-dependency samples only for group $\{2,3\}$, with the within-group probability of $0.4$.
Group $\{0,1\}$ appears exclusively in hard training samples, so approximately $20\%$ of the full training set has short dependencies.
We compare group-wise test accuracy against the hard-only and both-groups conditions from the preceding section.

\begin{figure}[t]
    \centering
    \includegraphics[width=\linewidth]{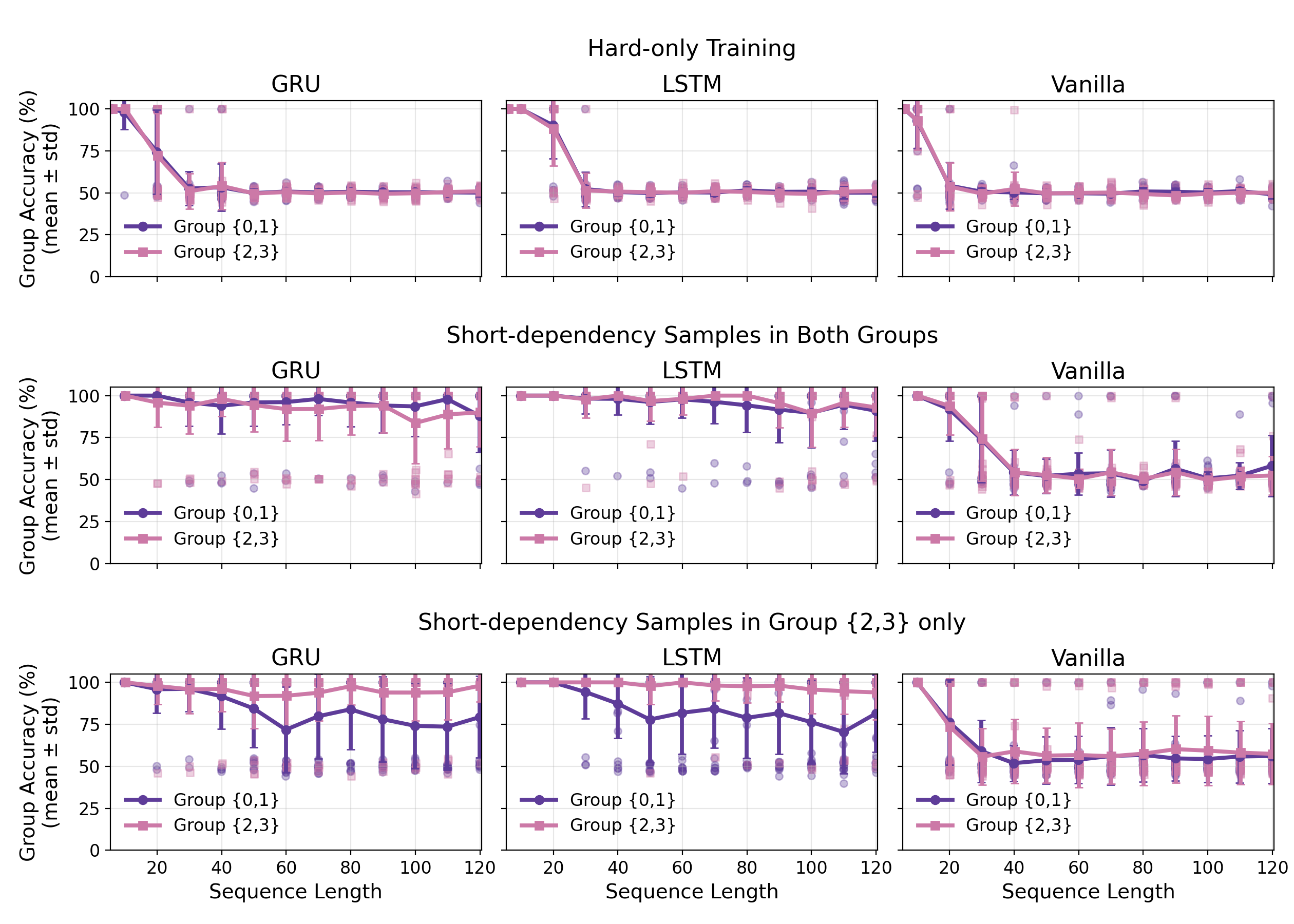}
    \caption{
    \textbf{Short-dependency training on one token group partially
    transfers to the other.}
    Columns show GRU, LSTM, and vanilla RNN results on multi-class XOR.
    Rows compare hard-only training, short-dependency samples in both groups, and short-dependency samples in group $\{2,3\}$ only.
    Short-dependency samples constitute $20\%$ of each group in the middle row and $40\%$ of group $\{2,3\}$ in the bottom row. They place both informative tokens within the final ten positions.
    Purple and pink show test accuracy for groups $\{0,1\}$ and $\{2,3\}$, respectively.
    All test samples place the informative tokens at the sequence endpoints; accuracy is computed using the original four-class predictions, separately for each group.
    Identifying the group but guessing between its two labels yields $50\%$ accuracy.
    Lines show means over $25$ runs, error bars show standard deviations, and faint markers show individual runs.
    With support restricted to group $\{2,3\}$, GRUs and LSTMs also improve on group $\{0,1\}$, but less than when both groups receive support. Gate-learning generalizes beyond token and label boundaries.
    }
    \label{fig:mxor-group-transfer}
\end{figure}

~\fig{mxor-group-transfer} shows that the two groups behave similarly under hard-only training and when both receive short-dependency samples.
Restricting this support to group $\{2,3\}$ reveals a clear asymmetry in GRUs and LSTMs.
The supported group retains mean accuracy above approximately $90\%$ across the tested lengths.
The unsupported group also improves substantially over the hard-only baseline, reaching approximately $70$--$85\%$ on longer sequences.
However, it performs worse than the supported group and than when it receives its own short-dependency samples.
The vanilla RNN shows no comparable improvement, with both groups remaining close to $50\%$ on long sequences.

The improvement on group $\{0,1\}$ demonstrates transfer of the training benefit across distinct informative tokens and output labels.
A natural interpretation is that the two groups share part of the required memory behavior.
They use the same non-informative tokens, so learning to preserve stored information across these tokens can benefit both groups.
Writing the relevant information and mapping it to an output, however, still involve group-specific tokens and labels.
The results are therefore consistent with a partially shared write-or-copy policy, while the remaining performance gap shows that support for one group does not fully replace direct support for the other.

\subsection{Learning Dynamics and Gradient Propagation}
\label{sec:learning-dynamics}

We now examine how learning unfolds, using binary temporal XOR at $T=100$ to focus on a single informative-token group.
Binary XOR has only two informative tokens $\{0,1\}$, the label is their XOR, and non-informative tokens are sampled uniformly from $\{2,\ldots,9\}$.
We compare a failed hard-only GRU run with a successful mixed run in which $40\%$ of training samples have short dependencies.

\begin{figure}[t]
    \centering
    \begin{tabular}{@{}c@{}c@{}}
    \multicolumn{2}{@{}c@{}}{%
        \includegraphics[width=\linewidth]{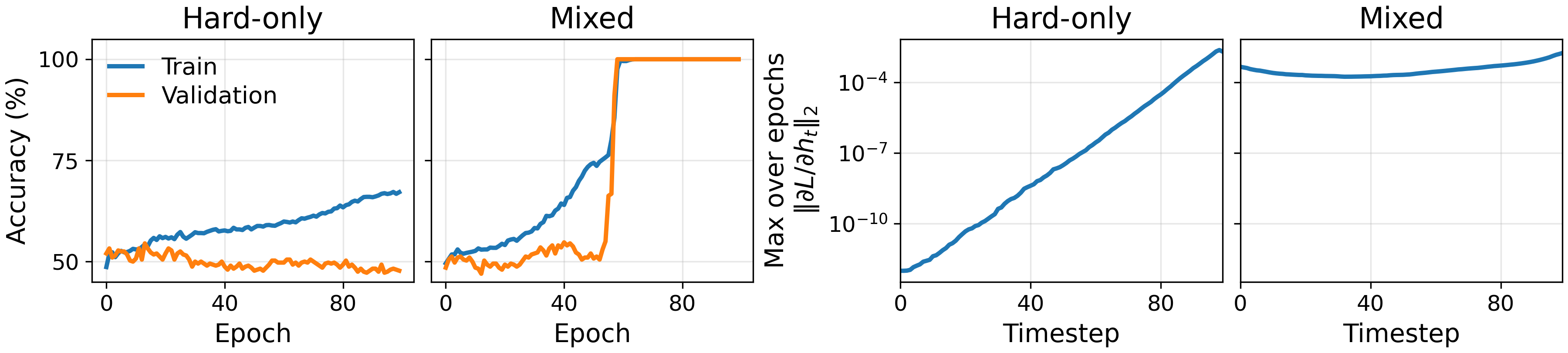}
    } \\
    \makebox[0.5\linewidth]{Training dynamics} &
    \makebox[0.5\linewidth]{Gradient propagation}
    \end{tabular}
    \caption{
    \textbf{Training dynamics and gradient propagation for two GRU runs on binary XOR at $T=100$.}
    The left pair shows training and validation accuracy; the right pair shows the maximum recorded hidden-state gradient norm at each sequence position.
    Hard-only samples place informative tokens between a long-range uninformative token sequence; mixed training places them within the final ten positions in $40\%$ of training samples.
    Validation uses hard samples exclusively, where chance accuracy is $50\%$.
    Hard-only validation accuracy remains near chance, whereas the mixed run reaches $100\%$ around epoch $60$, which is when the Gate-RNN is successfully trained and the Main-RNN can exploit the long-range samples for training.
    Early-position gradient maxima are approximately $10^{-12}$ under hard-only training and $10^{-4}$--$10^{-3}$ in the mixed run.
    Successful learning is therefore associated with stronger gradient signals reaching early recurrent states.
    }
    \label{fig:txor-gru-training}
    \label{fig:txor-gru-grad-over-t}
\end{figure}

The left pair in~\fig{txor-gru-training} reveals two distinct learning trajectories.
Under hard-only training, training accuracy gradually rises to approximately $70\%$, while validation accuracy remains near $50\%$, consistent with overfitting.
With short-dependency samples, training accuracy first improves while hard-validation accuracy remains near chance. This initial phase is consistent with the Gate-RNN needing to learn the gates based on the short-range data.
Around epochs $55$--$60$, both accuracies rapidly reach $100\%$, which is where the Gate-RNN has learned what to pass through, and the Main-RNN can start to take advantage of the long-range training samples. 

The right pair in~\fig{txor-gru-grad-over-t} shows a corresponding difference in gradient propagation.
Under hard-only training, the maximum recorded gradient norm falls by roughly nine orders of magnitude from the final positions to the beginning of the sequence.
In the successful mixed run, these maxima remain within roughly one order of magnitude across positions.
Early recurrent states thus receive much stronger gradients at some point during successful training.

These observations are consistent with the gate-learning account in Section~\ref{sec:gate-learning-implications}: Short-dependency samples offer shorter paths from informative inputs to the loss, allowing shared gates to be learned before the long-range task is solved. Learned gates then enable gradients to reach early positions.
Additional analyses of gate values and hidden-state updates are provided in Appendix~\ref{app:additional-probes}.

\subsection{Extension to Modern Gated RNNs and State-Space Models}
\label{sec:additional-architectures}

We next examine whether the gate-learning problem and the benefit of short-dependency samples extend to more general gating: minGRU and minLSTM, and compare their test accuracy with Mamba and an sLSTM-based xLSTM on binary XOR. LSTM serves as a reference to the preceding experiments.
We compare hard-only and mixed training for all five architectures.
In mixed training, $40\%$ of training samples use short-dependency placement.

\begin{figure}[t]
    \centering
    \includegraphics[width=\linewidth]{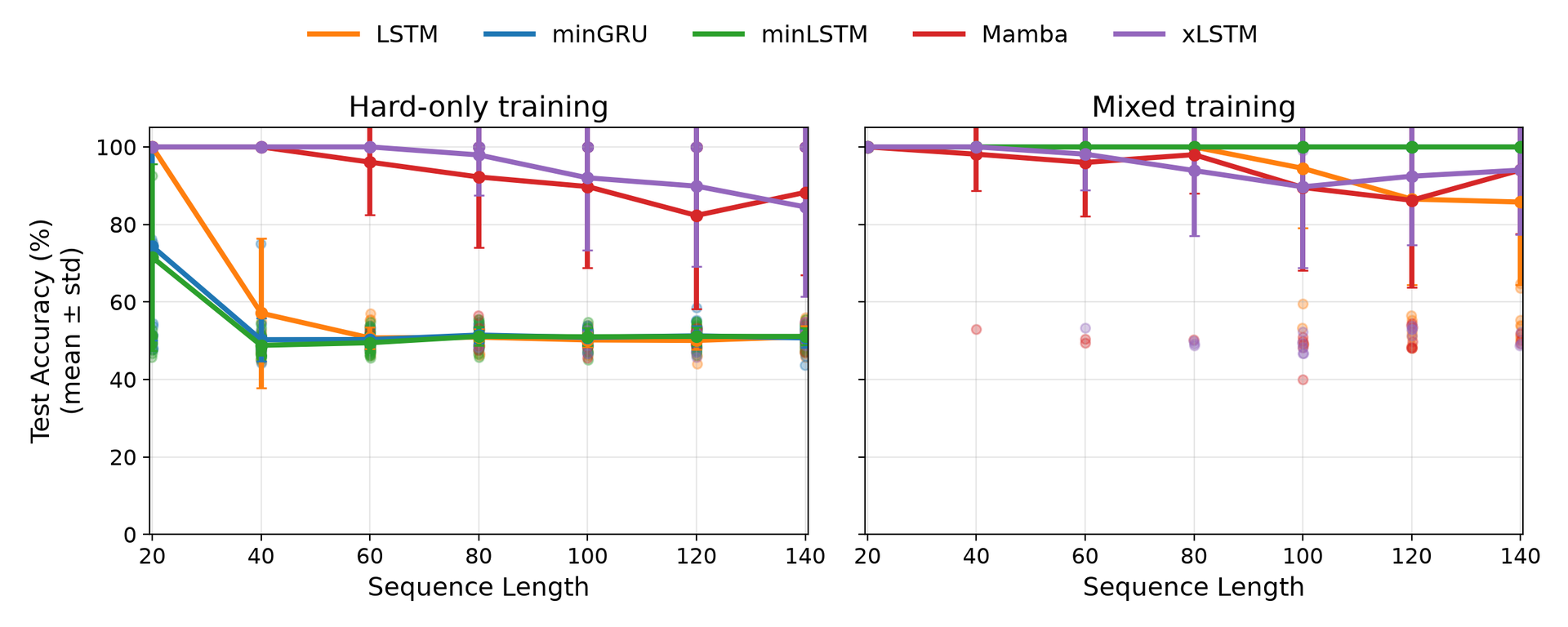}
    \caption{
    \textbf{Test accuracy on binary XOR for gated models and modern architectures.}
    Left: all models are trained with informative tokens at the sequence endpoints.
    Right: $40\%$ of training samples place both informative tokens within the final ten positions.
    All validation and test samples use endpoint placement.
    Lines show means over 25 runs, error bars show standard deviations, and faint markers show individual runs.
    With short-dependency samples, minGRU and minLSTM reach $100\%$ across all tested lengths; their curves overlap. LSTM also improves substantially.
    Thus, on this task, training support allows gated models to match or exceed the accuracy of Mamba and xLSTM.
    }
    \label{fig:txor-modern-comparison}
\end{figure}

Under hard-only training, LSTM, minGRU, and minLSTM approach chance accuracy as sequence length increases (~\fig{txor-modern-comparison}).
Mamba and xLSTM perform substantially better, maintaining mean accuracies above $80\%$ throughout the tested range, although some runs fail at longer lengths.

Introducing short-dependency samples changes this comparison.
Both minGRU and minLSTM achieve $100\%$ test accuracy in all runs at every tested length.
LSTM also improves markedly, reaching perfect accuracy through $T=80$ and approximately $85$--$95\%$ at longer lengths.

These results show that the gate-learning problem and the benefit of short-dependency samples extend to the more general gating in minGRU and minLSTM.
With training support, these gated RNN models match or exceed the performance of Mamba and xLSTM trained on the same mixed data.

\section{Discussion}
We study how gated recurrent networks learn to use their memory paths.
Our analysis shows that recurrent computation can attenuate gate-learning signals, even for input-only gates.
In controlled tasks, short-dependency samples help long-range learning in GRUs and LSTMs, with partial transfer across informative tokens and output labels.
The benefit extends to minGRU and minLSTM, which match or exceed the test accuracy of Mamba and xLSTM.

We identify several limitations.
The probes reveal changes in gradient propagation and selective updates, but their causal roles in learning long-range solutions are not isolated.
Interventions on learned gate functions could clarify which behaviors account for success and partial transfer.
It also remains unclear why Mamba and xLSTM learn more reliably from hard samples under the tested configurations.
A further challenge is to identify or introduce comparable short-dependency support in large-scale models trained on real-world data, such as language and images.

These findings highlight the dependency structure of training data as a practical way to improve long-range learning within existing recurrent architectures.
Exposing the same task over shorter dependencies can help shared gates acquire memory behavior that remains useful over longer intervals.
Effective gated memory therefore depends on both the available memory mechanism and the training signals that teach the model to use it.


\subsection*{AI Use Statement}

The original problem setting and hypothesis formulation were done without AI. We used ChatGPT (OpenAI, GPT-6 Astra) to assist with designing initial experiments, interpreting results, and developing and checking the mathematical derivations in Section~\ref{sec:gate-learning}.
We also used it to help implement and debug experimental code, including procedural data generation, parameter sweeps, and the code for figure plotting.
Additional uses included literature search and getting a very first draft of sections. A complete and careful revision of the manuscript's organization and of all wording was done manually by the authors.
The authors fully validated the mathematical arguments, cited sources, and experimental code.
The authors take full responsibility for the final manuscript, including its claims, results, references, and code.

\subsection*{Reproducibility Statement}

Section~\ref{sec:gate-learning} presents the mathematical derivations and their assumptions.
Section~\ref{sec:experiments} describes the synthetic tasks, data splits, training conditions, and evaluation protocol.
Appendix~\ref{app:model-configurations} specifies the model implementations and configurations, while Appendix~\ref{app:optimization-evaluation} documents optimization, stopping criteria, checkpoint selection, and the model and data seeds used in the repeated runs.
The diagnostic measurements are described in Section~\ref{sec:learning-dynamics} and Appendix~\ref{app:additional-probes}.

\clearpage
\bibliography{references}
\bibliographystyle{iclr2027_conference}

\clearpage
\appendix
\renewcommand*{\theHfigure}{\thesection.\arabic{figure}}
\setcounter{figure}{0}
\renewcommand{\thefigure}{A.\arabic{figure}}

\section{Experimental Details}
\label{app:ExpA}
\subsection{Model Configurations}
\label{app:model-configurations}

All classifiers use a learned 8-dimensional token embedding, one recurrent layer or sequence-model block, and a linear classification head applied to its output at the final timestep.
The head has two outputs for binary XOR and four for multi-class XOR.
Recurrent states are reset for every sequence.
We choose small configurations based on parameter count and recurrent-state size, with the resulting widths and parameter counts reported in Table~\ref{tab:app-model-widths}.

\begin{table}[t]
\centering
\small
\begin{tabular}{llrrr}
\hline
Experiment & Model & Hard-only width & Mixed width & \#Params \\
\hline
Basic recurrent models & Vanilla RNN & $10$ & $24$ & $366 / 1010$ \\
 & GRU & $10$ & $10$ & $766$ \\
 & LSTM & $10$ & $10$ & $966$ \\
\hline
Additional architectures & LSTM & $15$ & $15$ & $1676$ \\
 & minGRU & $20$ & $20$ & $546$ \\
 & minLSTM & $20$ & $20$ & $726$ \\
 & Mamba & $10$ & $10$ & $776$ \\
 & xLSTM (sLSTM) & $10$ & $10$ & $1576$ \\
\hline
\end{tabular}
\caption{Hidden or model widths.
Parameter counts are reported for binary XOR and include the embedding, backbone, and classifier.
The two counts for the vanilla RNN correspond to hard-only and mixed training, respectively.
For multi-class XOR, the corresponding counts are $500/1172$ for
the vanilla RNN, $900$ for GRU, and $1100$ for LSTM.
Mamba additionally uses a state dimension of $2$ per channel.}
\label{tab:app-model-widths}
\end{table}

\paragraph{Vanilla RNN, GRU, and LSTM.}
We use explicit sequential implementations of the standard recurrent updates, with a $\tanh$ activation for the vanilla RNN.
Recurrent weights and biases are initialized uniformly in $[-1/\sqrt{H},1/\sqrt{H}]$, where $H$ is the hidden width.
GRU and LSTM gate biases use this same initialization.
The initial hidden states, and the LSTM cell state, are zero.
These models use no dropout or normalization layers.

\paragraph{minGRU and minLSTM.}
Both models use input-only gates and linear input-only candidates.
minGRU combines its previous state and candidate using complementary sigmoid gate weights.
minLSTM normalizes its sigmoid forget and input gates by their sum; the implementation computes this normalization in the log domain for numerical stability.
Both use zero initial states and the same uniform parameter initialization as the basic recurrent models.

\paragraph{Mamba.}
We use a small sequential implementation of a single Mamba-1 block.
The embedding is projected to model width $10$, followed by RMS normalization, a selective state-space block, and a residual connection.
The state dimension per channel is $2$, the causal depthwise-convolution kernel size is $4$, the expansion factor is $1$, and the time-step projection rank is $1$.
The block includes the direct state-space skip connection and SiLU output gating.
The time-step initialization uses the range $[10^{-3},10^{-1}]$.

\paragraph{xLSTM.}
We use the official xLSTM implementation with one sLSTM block, model width $10$, and one head.
The input embedding is linearly projected to the model width.
The causal convolution has kernel size $4$, and the feedforward sublayer uses GELU with projection factor $1.3$ and width rounding to a multiple of $1$.
The configuration enables post-block normalization and uses zero dropout, zero-initialized recurrent weights, and the implementation's \texttt{powerlaw\_blockdependent} bias initialization.
We use its \texttt{vanilla} backend in float32, with automatic mixed precision and recurrent-gradient cutting disabled.

\subsection{Optimization and Evaluation}
\label{app:optimization-evaluation}

All models minimize cross-entropy loss using AdamW with batch size $500$, weight decay $10^{-4}$, and $(\beta_1,\beta_2)=(0.9,0.95)$.
The learning rate is $5\times10^{-3}$ for binary XOR, including the additional architecture comparison, and $10^{-2}$ for multi-class XOR.
Training uses full backpropagation through the sequence, without gradient clipping or a learning-rate schedule.
All trainable parameters belong to the same optimizer parameter group.
Training lasts at most $500$ epochs and stops early after $20$ consecutive epochs with $100\%$ validation accuracy.
We evaluate validation accuracy after every epoch and retain the checkpoint with the highest accuracy, breaking ties by lower validation loss.
Test results are computed after restoring this checkpoint.

\paragraph{Repeated runs.}
Each sweep setting uses all combinations of model seeds and data seeds from $\{11,22,33,44,55\}$, giving $25$ runs per model, sequence length, and training condition.
This protocol also applies to the additional architecture comparison and the suffix-window sweep.
The model seed sets the PyTorch random seed.
For data seed $s$, separate generators use seeds $s+10$, $s+15$, and $s+20$ for training, validation, and test generation, and $s+30$ for training-data shuffling.
We report the mean and standard deviation of test accuracy across runs; faint markers in the figures show individual runs.
Multi-class group accuracies are computed from the original four-class predictions, restricted to examples from the corresponding informative-token group.

\setcounter{figure}{0}
\renewcommand{\thefigure}{B.\arabic{figure}}

\section{Additional Diagnostic Probes}
\label{app:additional-probes}
We complement the GRU training and gradient analysis in the main text with token-conditioned gate and hidden-state probes, followed by corresponding diagnostics for LSTM and the vanilla RNN.
All examples use binary XOR at $T=100$.
Mixed training includes short-dependency placement in $40\%$ of training samples, while validation uses hard samples exclusively.
The figures illustrate individual runs rather than averages over the seed sweeps.

\paragraph{Probe definitions.}
Every three epochs, we evaluate one shuffled training batch with the current model in evaluation mode.
For each token category---non-informative tokens, token $0$, and token $1$---we average gate values over hidden dimensions and all occurrences of that category in the batch.
Hidden-state changes are measured as $\|h_t-h_{t-1}\|_2$ and averaged over the same occurrences, taking the initial state to be zero.

\subsection{GRU: Gate Values and Hidden-State Changes}
\label{app:gru-probes}

\begin{figure}[htbp]
\centering
\includegraphics[width=0.95\linewidth]{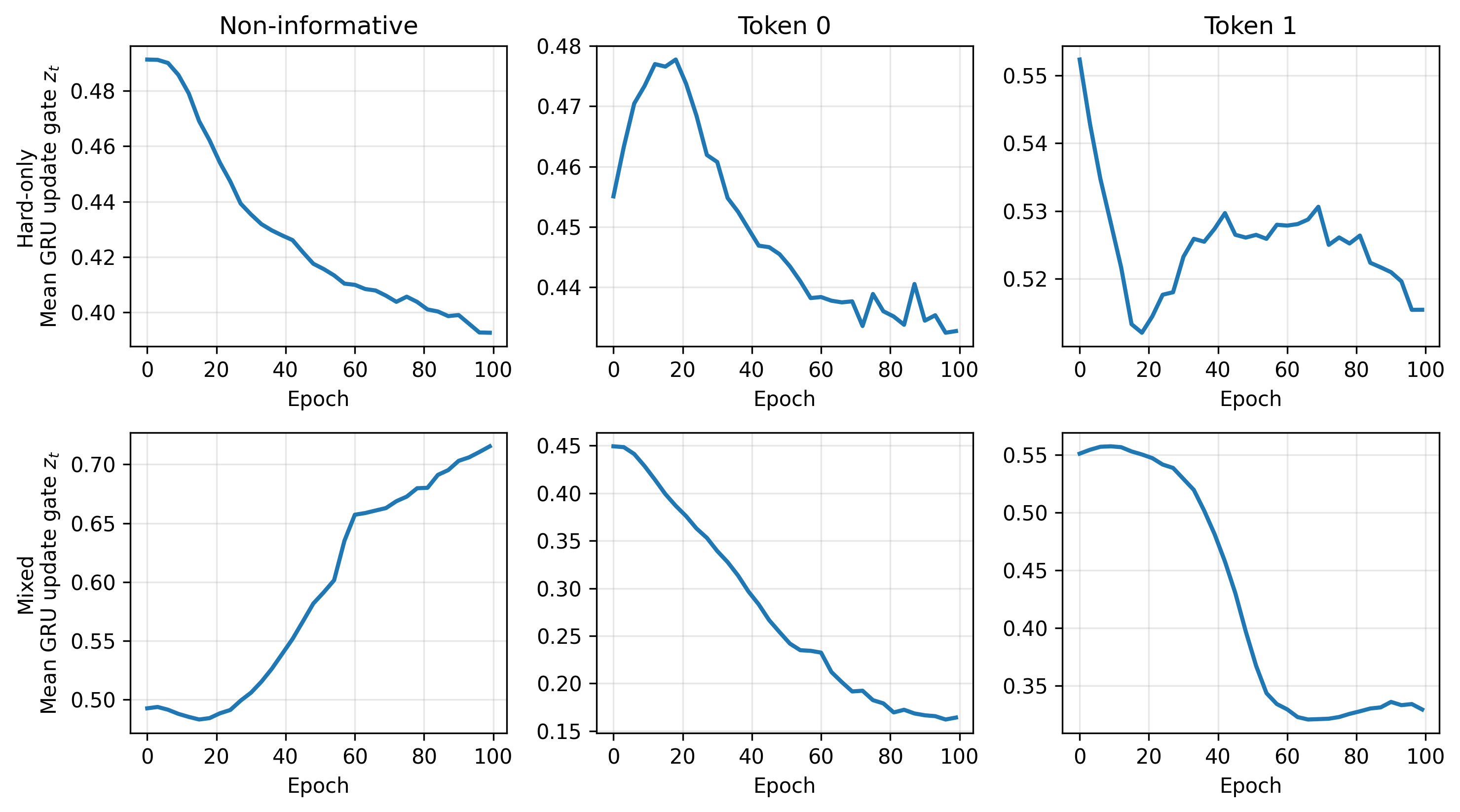}
\caption{\textbf{Token-conditioned GRU gate values.}
Columns show non-informative tokens, token $0$, and token $1$; rows show hard-only and mixed training.
Curves average $z_t$ over hidden dimensions and token occurrences in the probed batch.
Larger $z_t$ gives greater weight to the previous hidden state.
The mixed run develops higher retention weights for non-informative tokens and lower weights for informative tokens.
}
\label{fig:app-gru-gates}
\end{figure}

These probes extend the GRU trajectories shown in the main text.
Our implementation uses
\begin{equation}
    h_t=z_t\odot h_{t-1}+(1-z_t)\odot\widetilde{h}_t,
\end{equation}
so larger $z_t$ places more weight on the previous state.
~\fig{app-gru-gates} shows that the successful mixed run develops different gate responses to informative and non-informative tokens.
In the hard-only setting, the gate values for informative and non-informative tokens are not clearly separated.
In the mixed setting, the update gate becomes larger for non-informative tokens and smaller for informative tokens.
Thus, the successful mixed run learns a token-dependent gate policy that copies across non-informative tokens and writes when an informative token appears.

\begin{figure}[htbp]
\centering
\includegraphics[width=0.95\linewidth]{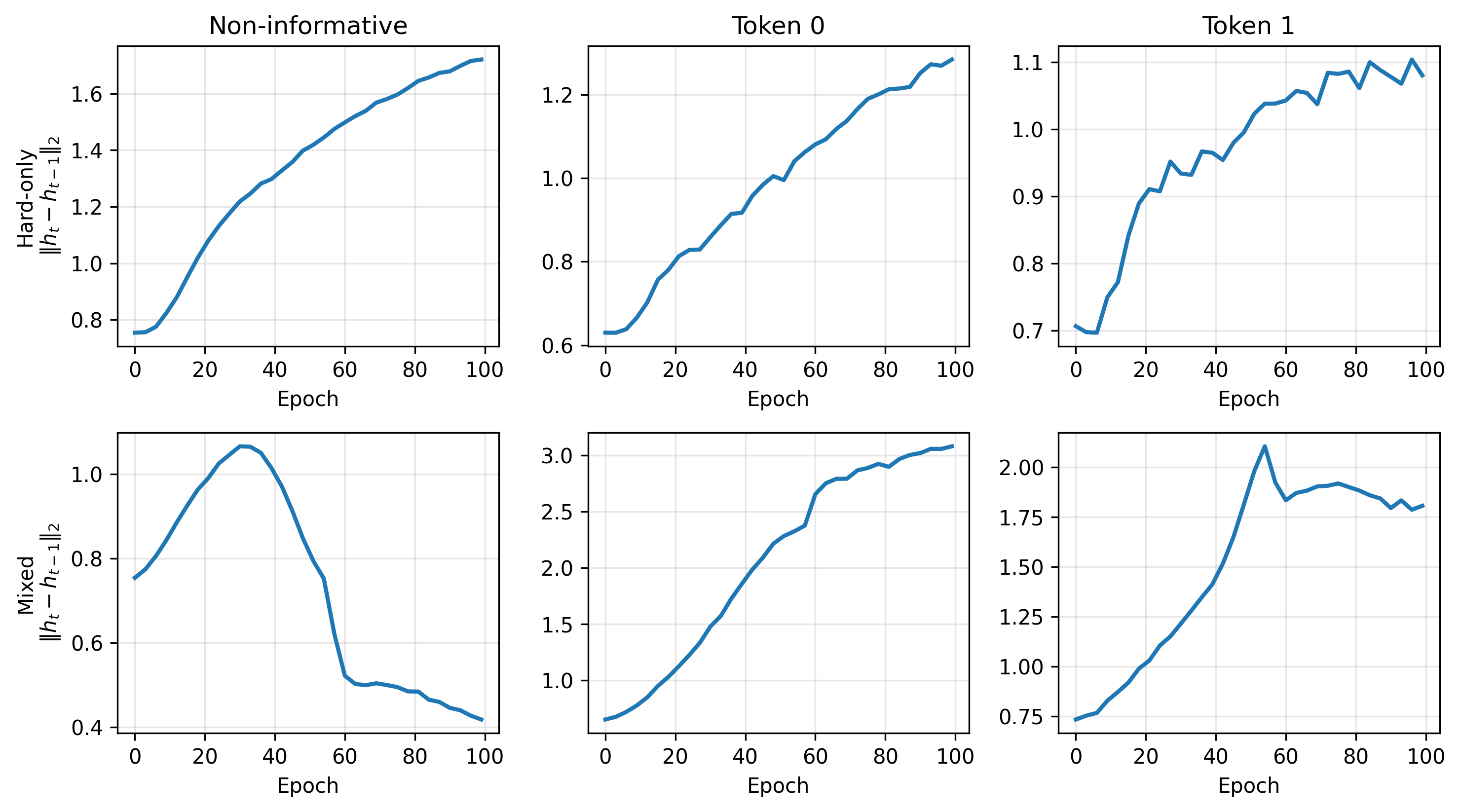}
\caption{\textbf{Token-conditioned GRU hidden-state changes.}
The layout matches ~\fig{app-gru-gates}.
Curves show the mean $\|h_t-h_{t-1}\|_2$ over occurrences of each token category.
The hard-only run shows increasing changes in all three categories.
In the mixed run, changes at non-informative tokens decrease, while informative tokens produce larger updates.
}
\label{fig:app-gru-dh}
\end{figure}

Gate averages describe only one component of the update.
~\fig{app-gru-dh} therefore measures the resulting hidden-state changes directly.
Under hard-only training, update magnitudes increase for all three token categories.
In the mixed run, changes caused by non-informative tokens initially increase, then fall substantially around the transition to successful hard-validation performance and continue to decrease.
Updates at informative tokens remain larger.
Together, the two probes are consistent with learning to preserve the state across non-informative inputs while updating it at informative inputs.

\FloatBarrier
\subsection{LSTM: Additional Gated-Model Diagnostics}
\label{app:lstm-probes}

\begin{figure}[htbp]
\centering
\includegraphics[width=0.95\linewidth]{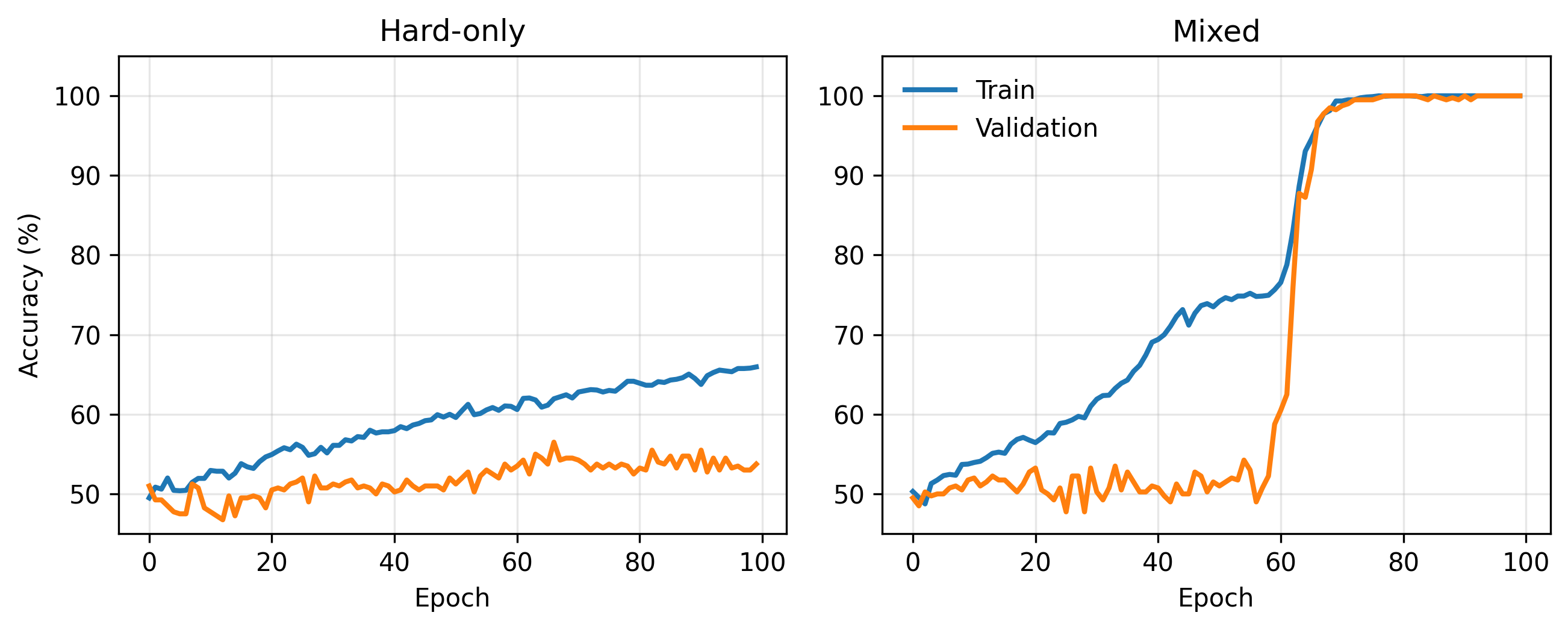}
\caption{\textbf{LSTM training and validation trajectories.}
Left: hard-only training. Right: mixed training.
Validation always uses hard samples, with chance accuracy of $50\%$.
The mixed run reaches high hard-validation accuracy after an initial period of training improvement; the hard-only run remains near chance on validation.}
\label{fig:app-lstm-training}
\end{figure}

\begin{figure}[htbp]
\centering
\includegraphics[width=0.95\linewidth]{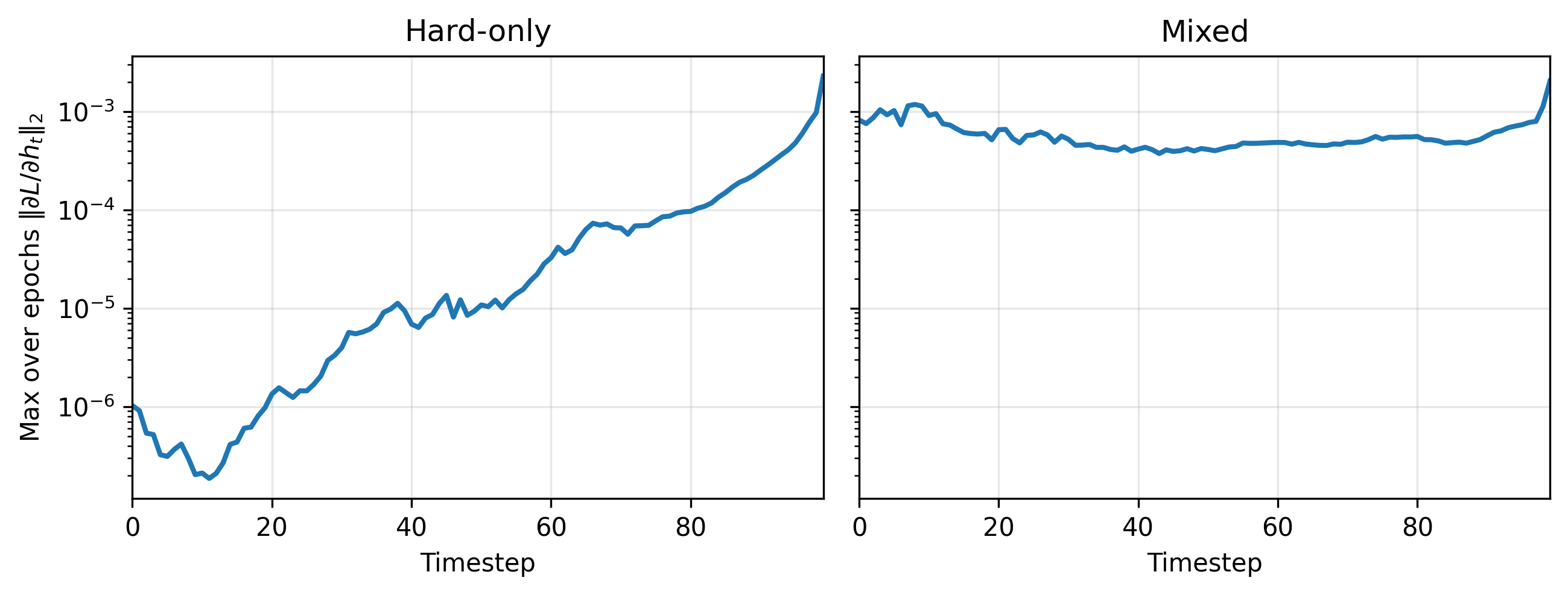}
\caption{\textbf{Maximum recorded LSTM hidden-state gradients.}
At each sequence position, curves show the maximum batch-averaged $\|\partial L/\partial h_t\|_2$ across probe epochs, on a logarithmic scale.
Early-position gradients are substantially larger in the mixed run than in the hard-only run.
}
\label{fig:app-lstm-grad}
\end{figure}

The LSTM trajectories provide a second example of successful learning with short-dependency support.
In ~\fig{app-lstm-training}, hard-only validation accuracy remains near chance, whereas mixed training produces a sharp rise in both training and hard-validation accuracy.
~\fig{app-lstm-grad} shows stronger gradients reaching early hidden states during the mixed run.

\begin{figure}[htbp]
\centering
\includegraphics[width=0.95\linewidth]{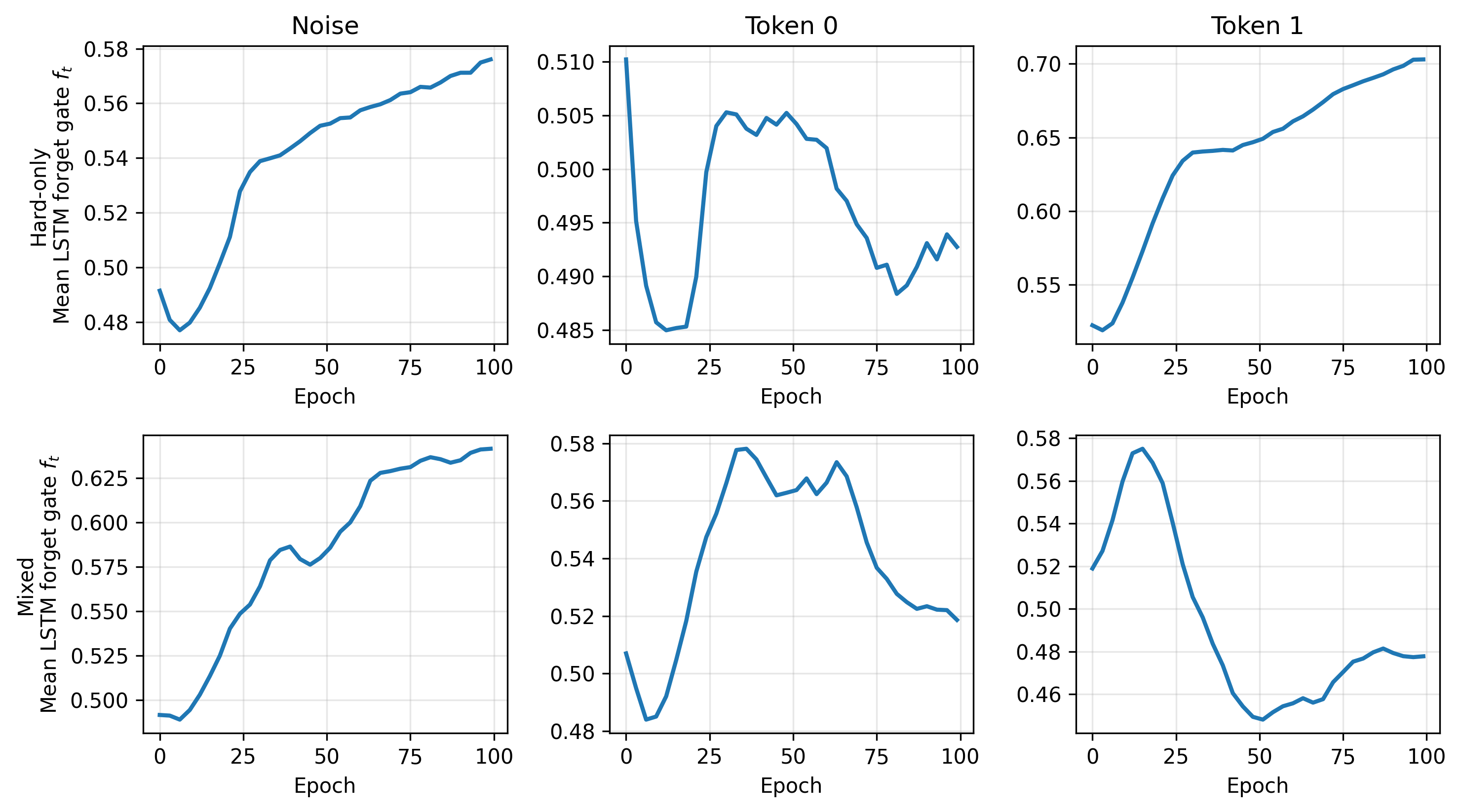}
\caption{\textbf{Token-conditioned LSTM forget-gate values.}
Columns show non-informative tokens, token $0$, and token $1$; rows show hard-only and mixed training.
Curves average $f_t$ over hidden dimensions and token occurrences.
Larger values give greater weight to the previous cell state.
The mixed run shows increasing retention weights for non-informative tokens, while the complete state update also depends on the other gates and the candidate.}
\label{fig:app-lstm-forget}
\end{figure}

\begin{figure}[htbp]
\centering
\includegraphics[width=0.95\linewidth]{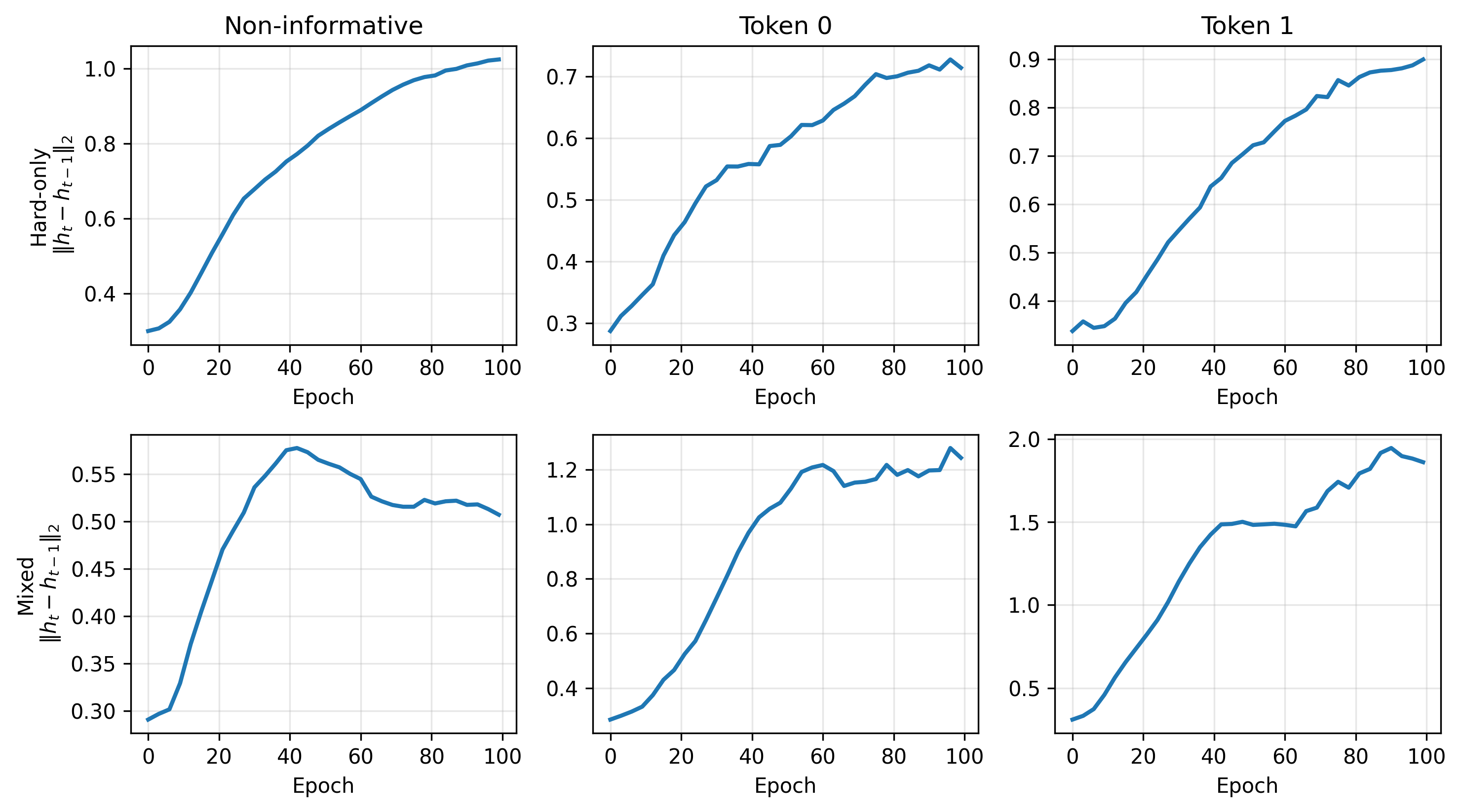}
\caption{\textbf{Token-conditioned LSTM hidden-state changes.}
The layout matches ~\fig{app-lstm-forget}.
Curves show the mean $\|h_t-h_{t-1}\|_2$ for each token category.
In the mixed run, updates become more selective, with smaller changes at non-informative tokens relative to informative tokens.
}
\label{fig:app-lstm-dh}
\end{figure}

~\fig{app-lstm-forget} examines the forget gate $f_t$, which weights the previous cell state in
\begin{equation}
    c_t=f_t\odot c_{t-1}+i_t\odot\widetilde{c}_t.
\end{equation}
The mixed run shows increasing forget-gate values for non-informative tokens.
However, cell-state changes also depend on the input gate and candidate, while the hidden state additionally depends on the output gate.
The forget-gate average alone therefore does not determine whether the state is preserved.
The effective hidden-state changes in ~\fig{app-lstm-dh} provide complementary evidence: the mixed run develops smaller updates at non-informative tokens relative to informative tokens.
This is consistent with the selective-update behavior observed in the GRU.

\FloatBarrier
\subsection{Vanilla RNN: A Non-Gated Comparison}
\label{app:vanilla-probes}

\begin{figure}[htbp]
\centering
\includegraphics[width=0.95\linewidth]{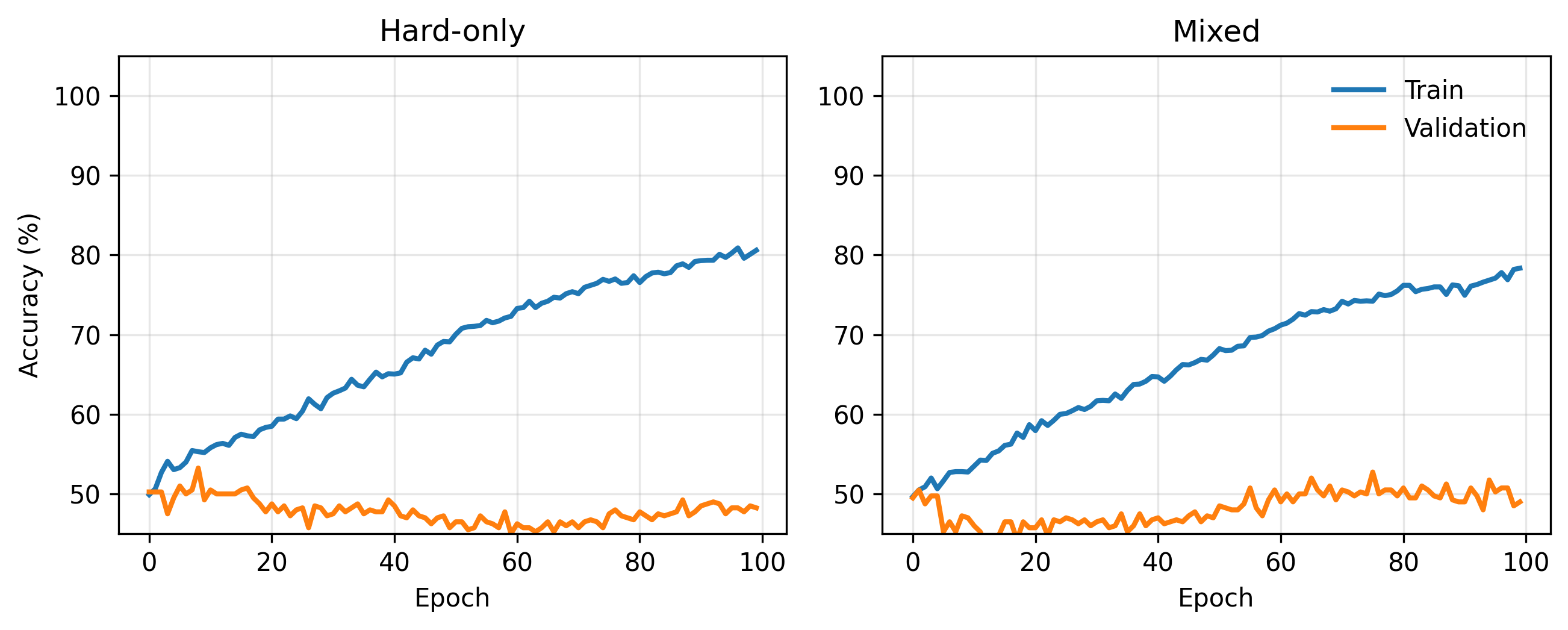}
\caption{\textbf{Vanilla RNN training and validation trajectories.}
Left: hard-only training. Right: mixed training.
Training accuracy increases in both runs, while accuracy on hard validation samples remains near the $50\%$ chance level.}
\label{fig:appendix-vanilla-training}
\end{figure}

\begin{figure}[htbp]
\centering
\includegraphics[width=0.95\linewidth]{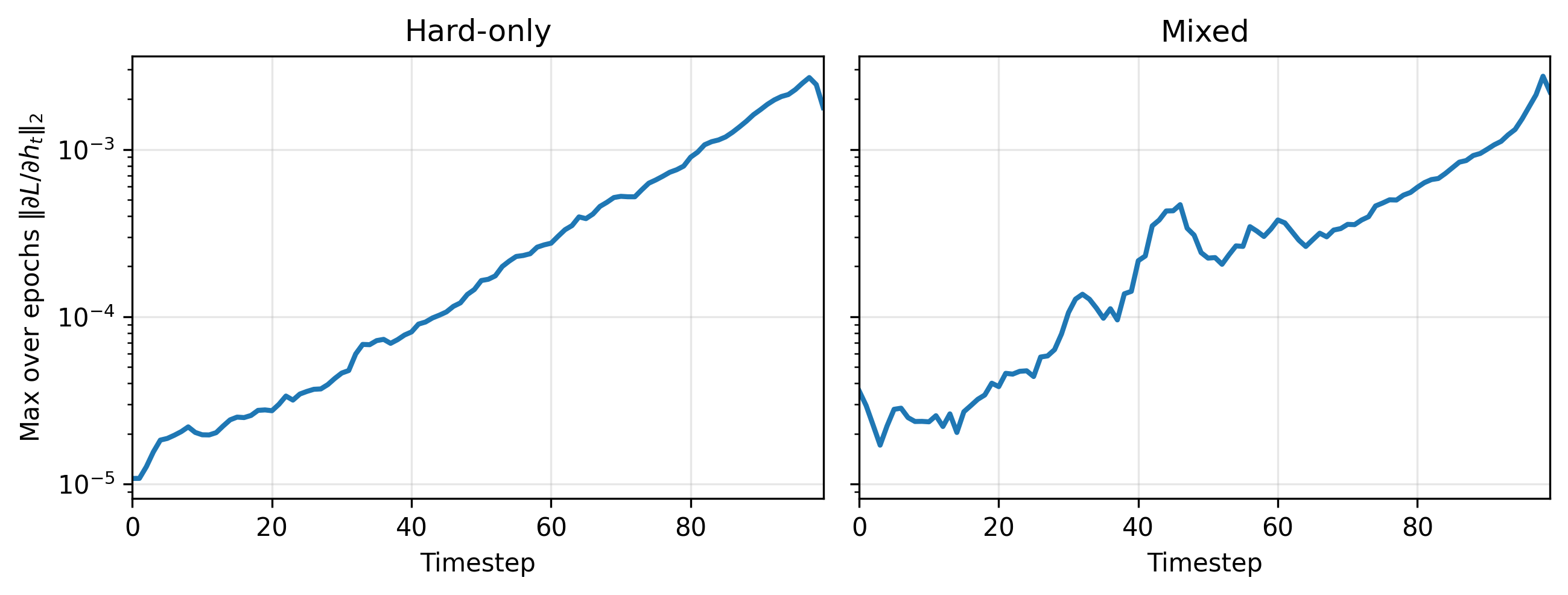}
\caption{\textbf{Maximum recorded vanilla RNN hidden-state gradients.}
Curves show the maximum batch-averaged $\|\partial L/\partial h_t\|_2$ at each position across probe epochs, on a logarithmic scale.
Under both training conditions, gradients are much smaller at early positions than near the final prediction.}
\label{fig:appendix-vanilla-gradient}
\end{figure}

\begin{figure}[htbp]
\centering
\includegraphics[width=0.95\linewidth]{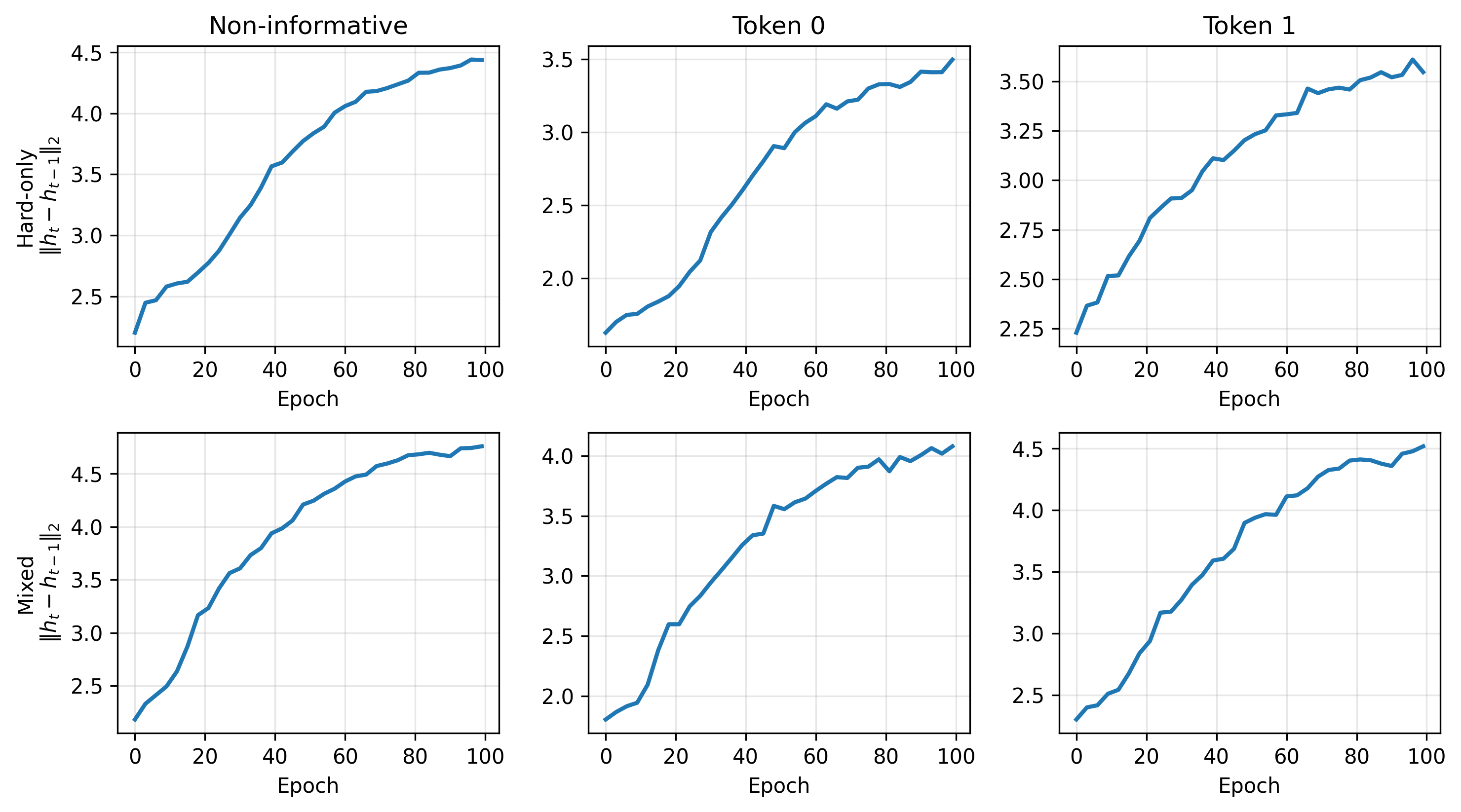}
\caption{\textbf{Token-conditioned vanilla RNN hidden-state changes.}
Columns show non-informative tokens, token $0$, and token $1$; rows show hard-only and mixed training.
Curves show the mean $\|h_t-h_{t-1}\|_2$ over occurrences of each token category.
Non-informative tokens continue to produce substantial updates, including in the mixed run.}
\label{fig:appendix-vanilla-dh}
\end{figure}

The vanilla RNN receives the same form of short-dependency support but does not reach high hard-validation accuracy in the illustrated runs (~\fig{appendix-vanilla-training}).
Its gradient profiles remain strongly concentrated near the end of the sequence under both training conditions (~\fig{appendix-vanilla-gradient}).
The token-conditioned state changes also show no comparable separation into small updates at non-informative tokens and large updates at informative tokens (~\fig{appendix-vanilla-dh}).
Non-informative inputs continue to produce substantial hidden-state changes during mixed training.

Across these illustrative runs, successful learning in GRU and LSTM accompanies stronger early-state gradients and more selective hidden-state updates.
The vanilla runs show neither the same validation improvement nor the same diagnostic pattern.
These observations support the proposed connection between short-dependency support and selective memory.

\FloatBarrier

\setcounter{figure}{0}
\renewcommand{\thefigure}{C.\arabic{figure}}

\section{Additional Experiments}
\label{app:additional-exp}
\subsection{Sensitivity to the Short-Dependency Window Size}
\begin{figure}[htbp]
  \centering
  \includegraphics[width=0.95\linewidth]{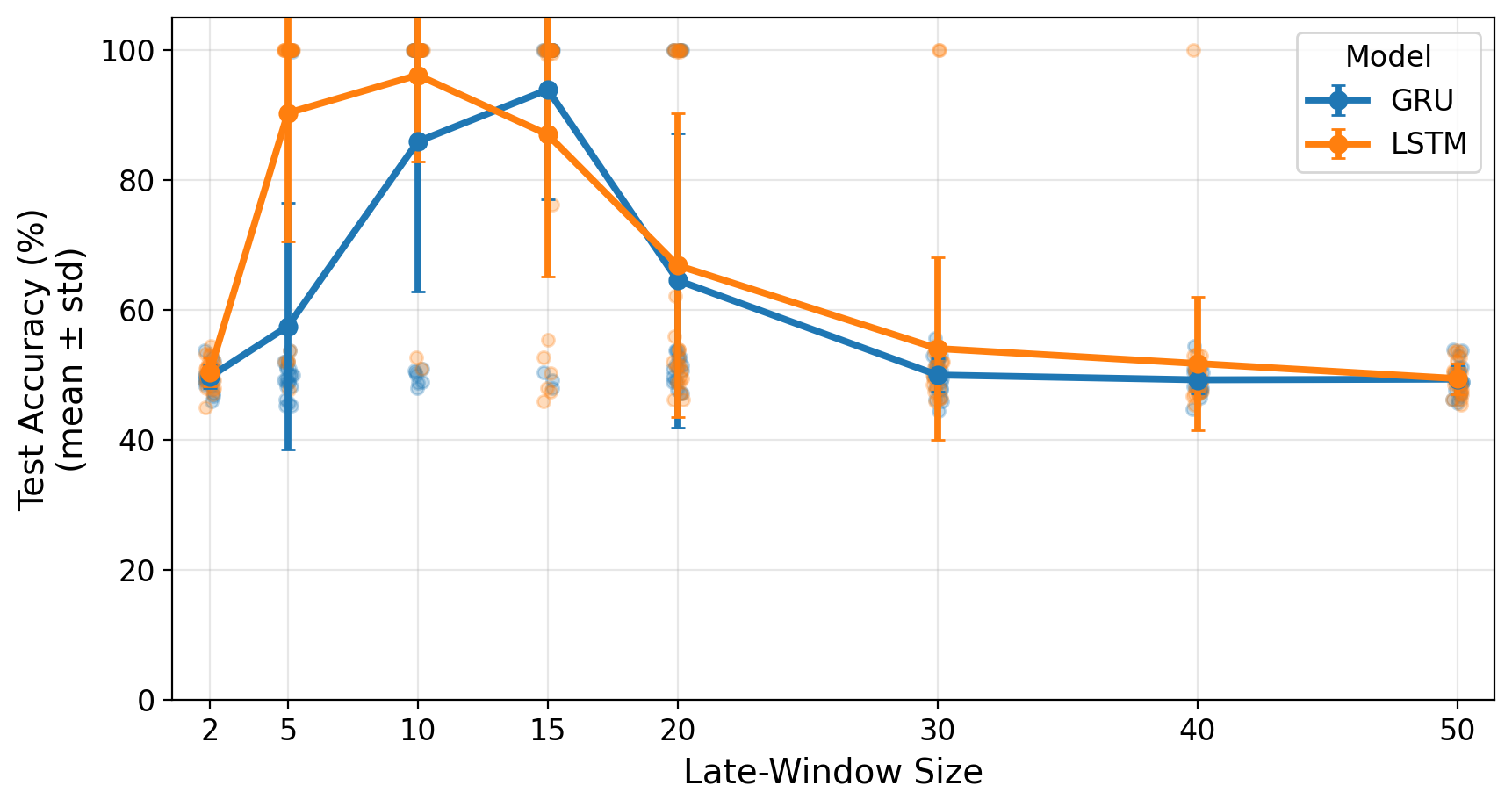}
  \caption{
  \textbf{Sensitivity to the short-dependency suffix-window size.}
  Sequence length is fixed to $T=100$, the short-dependency-sample proportion is $p_{\mathrm{easy}}=40\%$, and test samples are always hard.
  Very small windows and very large windows both reduce performance.
  Intermediate windows work best: they are short enough to provide usable gradient signals, but large enough to include non-informative tokens from which the model can learn when to copy state.
  The main experiments use suffix-window size $10$, which lies in this high-performing regime.
  }
  \label{fig:appendix-late-window}
\end{figure}

In the main experiments, short-dependency samples place the informative tokens inside a suffix window of size $10$.
To test whether this choice is special, we repeat the mixed binary temporal XOR experiment at fixed sequence length $T=100$ while varying the suffix-window size.
~\fig{appendix-late-window} shows an inverse-U pattern.
Very small windows provide limited non-informative context, while very large windows make the short-dependency samples less easy by increasing the dependency length again.
The best performance is obtained for intermediate window sizes, including the window size used in the main experiments.

\subsection{Symmetry Check for the Multi-Class XOR Setting}
\label{app:mxor-symmetry}

In the main multi-class XOR transfer experiment, short-dependency samples are added only to the $\{2,3\}$ token group.
Here we repeat the asymmetric experiment in the opposite direction: short-dependency samples are added only to the $\{0,1\}$ group, while the $\{2,3\}$ group remains hard-only during training.
All validation and test samples remain hard samples for both groups.

\begin{figure}[htbp]
\centering
\includegraphics[width=0.95\linewidth]{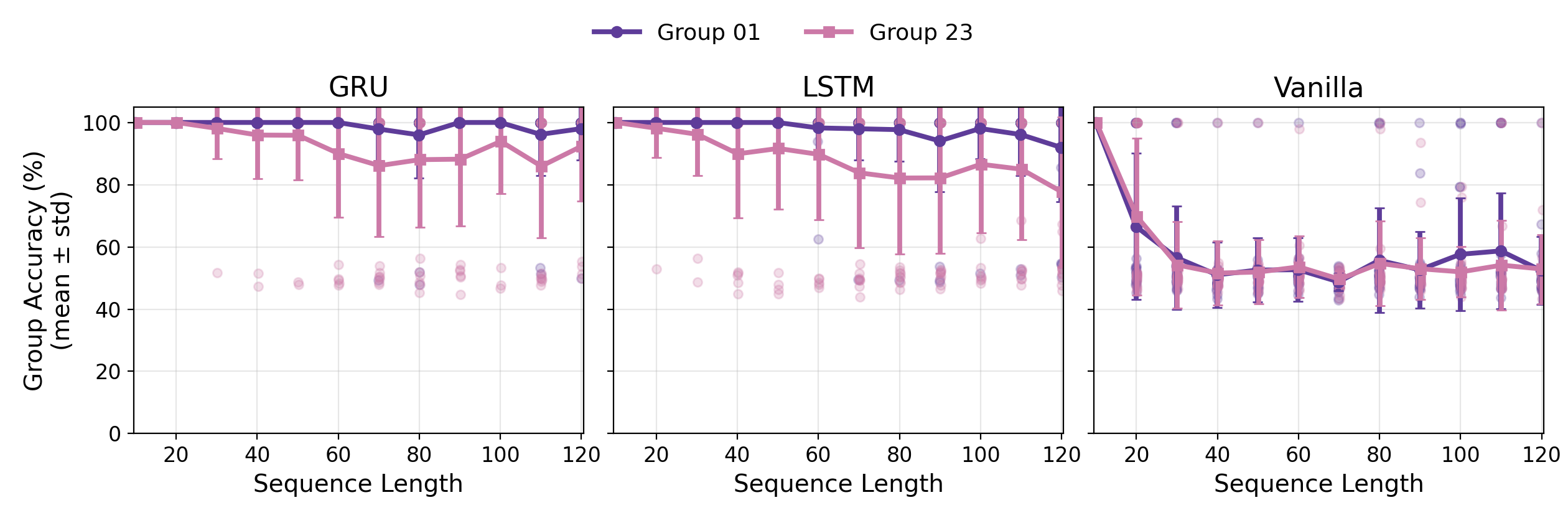}
\caption{
\textbf{Group-wise accuracy when short-dependency samples are added only to the $\{0,1\}$ group.}
The purple curve shows the treated $\{0,1\}$ group, and the pink curve shows the untreated $\{2,3\}$ group.
Lines show mean group accuracy, error bars show standard deviation, and faint markers show individual runs.
For GRUs and LSTMs, the treated group remains highly accurate, while the untreated group also improves over the hard-only baseline.
The untreated group is less stable than the treated group and remains below the fully treated reference condition from the main text.
Thus, the reverse asymmetric experiment supports the same conclusion: the learned behavior transfers beyond the treated token group, but the transfer is partial rather than complete.
}
\label{fig:app-mxor-mixed01-groups}
\end{figure}

~\fig{app-mxor-mixed01-groups} shows a qualitatively symmetric result to the setting in the main text.
The group-wise curves show that the treated $\{0,1\}$ group is learned reliably, while the untreated $\{2,3\}$ group also improves compared with the hard-only baseline.
However, the untreated group remains less stable than the treated group and does not match the fully treated reference condition from the main text.
This supports the interpretation that the benefit of short-dependency samples is not specific to the particular $\{2,3\}$ group used in the main experiment.
Instead, mixed training induces a partially reusable recurrent strategy: part of the behavior transfers across token groups, but direct short-dependency samples are still useful for learning how to write group-specific informative tokens.

\end{document}